\pdfoutput=1
\documentclass[11pt]{article}

\usepackage[final]{coling}

\usepackage{times}
\usepackage{latexsym}

\usepackage{amsmath}

\usepackage[T1]{fontenc}
\usepackage[utf8]{inputenc}

\usepackage{microtype}

\usepackage{inconsolata}

\usepackage{graphicx}
\title{Evaluating K-Fold Cross Validation for Transformer Based Symbolic Regression Models}

\author{%
Kaustubh Kislay \quad Shlok Singh \quad Soham Joshi \quad \textbf{Rohan Dutta} \\  \textbf{Jay Shim}\thanks{Senior Author} \quad \textbf{George Flint}\footnotemark[1] \quad \textbf{Kevin Zhu}\footnotemark[1]
\\
Algoverse AI Research\\
\texttt{kaustubh.kislay@gmail.com, kevin@algoverse.us}
}

\begin{document}
\maketitle
\begin{abstract}
Symbolic Regression remains an NP-Hard problem, with extensive research focusing on AI models for this task. Transformer models have shown promise in Symbolic Regression, but performance suffers with smaller datasets. We propose applying k-fold cross-validation to a transformer-based symbolic regression model trained on a significantly reduced dataset (15,000 data points, down from 500,000). This technique partitions the training data into multiple subsets (folds), iteratively training on some while validating on others. Our aim is to provide an estimate of model generalization and mitigate overfitting issues associated with smaller datasets. Results show that this process improves the model's output consistency and generalization by a relative improvement in validation loss of 53.31\%. Potentially enabling more efficient and accessible symbolic regression in resource-constrained environments. \end{abstract}

\section{Introduction}

Symbolic Regression (SR) is a machine learning (ML) technique used to uncover non-obvious relationships from complex datasets. Recent advancements in symbolic regression, including transformer-based models, have made it a powerful tool for complex data analysis. \citep{lacava2021symbolic} Similar to other ML models, the performance of SR is highly dependent on the size of the training dataset, with larger datasets generally leading to better outcomes compared to smaller ones. \citep{valipour2021symbolicgpt} Situations where training data is lacking due to resource constraining circumstances can therefore be a great limiter for these transformer-based SR models. \citep{valipour2021symbolicgpt} Enhancing the performance of SR models on smaller data sets is therefore essential to making advanced data analysis more accessible, allowing resource-constrained organizations to accelerate research across various fields, such as physics or finance. \citep{wang2020small} Such an approach would make transformer-based symbolic regression tasks more efficient and accessible.

Unlike traditional forms of regression, SR does not require a predefined form to build an equation. \citep{symbolic_regression_ieee} Instead, it can include arithmetic, logarithmic/exponential functions, trigonometric functions, and exponentiation in a generated equation. Additionally, although traditional regression models will find the structure of the equation before searching and applying parameters to a given structure, SR can find both the structure and parameters of an equation simultaneously.  

For example, genetic algorithms are a classical approach for conducting SR because they evolve a population of mathematical expressions through iterative selection, crossover, and mutation operations to find an optimal model that fits given data. \citep{symbolic_regression_ieee} Unlike these algorithms, SymbolicGPT, the model we experiment with, is classified as a Transformer model. \citep{valipour2021symbolicgpt} Transformer models have been pivotal in advancing tasks across various fields due to their ability to handle sequential data efficiently. \citep{vaswani2017attention} Its function consists of 3 stages: stage 1 is to use a T-net to get an order invariant embedding of the input dataset, stage 2 is to get the equation’s structure, or skeleton, using a GPT language model; and stage 3 is to optimize constant values to fill the "skeleton." \citep{valipour2021symbolicgpt} The training data, which is mainly synthetic, consists of sets of input and output data points along with the correct equation, which follows a predefined skeleton.  
These models have been proven to excel at generating fitting equations with robust and consistently low error values and high generalizability. \citep{valipour2021symbolicgpt} But this is only true given the dataset for which the SR model is trained is large enough. The generalizability of models cannot persist with an exceptionally small training dataset due to the dangers of overfitting, for example. In such cases, a form of model validation can manage the detriments of a small dataset.  \citep{raschka2020model}

K-fold cross-validation (KFCV), our chosen form of validation, is widely used to assess the model generalization, which is a model’s ability to perform well against new data. \citep{kohavi1995cross} It works by partitioning a given training dataset into multiple subsets called folds and iteratively training the model on complementary folds while validating the remaining ones. KFCV is thereby adapted to accept datasets for SR in JSON format to train, validate, and test on.  

In this work, we answer the question, "How does KFCV impact the generalizability of SymbolicGPT when trained on small datasets?" To do so, we investigate the application of KFCV on SymbolicGPT’s one-variable model trained with a dataset 97\% smaller than normal from 500,000 indices to 15,000 to assess its impact on model generalization with KFCV, the base model. \citep{wang2020small} Through our experiments, we contribute a reliable implementation of KFCV onto SymbolicGPT and show improvements in generalization/reduced overfitting by 53.31 percent. 

\section{Methods}

    The given training size for the 1 variable model configuration totaled --500,000 different indices, that being each input/output set represents one index, or that each line in the JSON file represents 1 index. We cut the total data down to \~15,000 indices, a 97\% decrease. 

For our implementation of K-Fold Cross Validation (KFCV), we used 5 folds, where k = 5. The dataset is split into k folds and is randomly sampled into these folds. The model is trained and validated k times, where each fold is used as the validation set exactly once. \citep{raschka2020model} During each iteration: 

\begin{itemize}
    \item \textbf{k-1 folds} are used for training the model.
    \item \textbf{1 fold} is used as the validation set to evaluate the model's performance.
\end{itemize}
 
\begin{figure}[t]
  \includegraphics[width=\columnwidth]{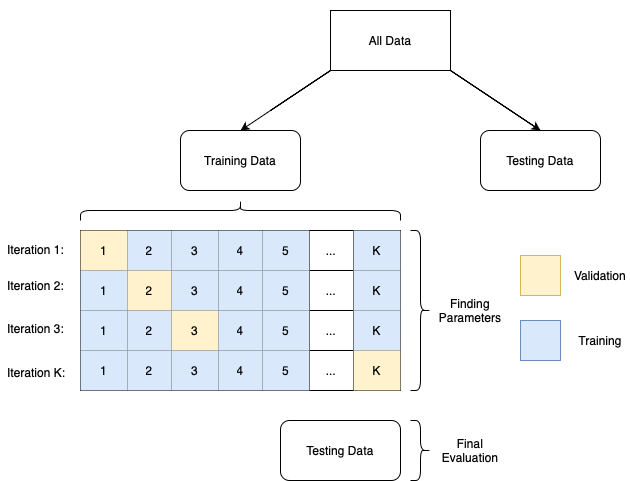}
  \caption{The following displays a diagram of K-Fold Cross Validation}
  \label{fig:experiments5}
\end{figure}

After training and validating the model k times (once for each fold), the training and validation error loss is taken from each iteration and averaged to find the average training loss and average validation loss. Model training hyperparameters are the same for both the applied model and the original model. The improvements are thereby calculated based on these loss values using equation 1 to find the relative improvement.  

\begin{equation}
  \left( \frac{\text{Old\_val\_loss} - \text{New\_val\_loss}}{\text{Old\_val\_loss}} \right) \times 100
\end{equation}

The original model utilizes an 80/20 train to test split but is additionally validated using testing data after each epoch. The loss values and overall performance of the old model are set to be a source of comparison to the applied model.

\section{Related Works}

\subsection{Don't Waste Your Time: Early Stopping Cross-Validation}
\citealp[]{bergman2024early} presents an approach to enhance model selection efficiency in automated machine learning systems by introducing early stopping in KFCV. This method addresses the computational overhead associated with KFCV, particularly in time-constrained model selection processes. Their work and ours both aim to improve model generalization and validation by leveraging CV, although with different focuses. While they investigate methods to reduce the computational cost associated with KFCV through early stopping, our work focuses on the benefits of KFCV in improving model performance across different categories of equations.

\subsection{Stratified cross-validation for unbiased and privacy-preserving federated learning} \citealp[]{bey2024fold} explores privacy-preserving techniques in federated learning settings and introduces a new method to avoid data leakage through stratified cross-validation (SCV). Similar to our goal of enhancing generalization, this work addresses concerns regarding improper model validation due to duplicate data across different institutions. However, while this paper’s approach focuses on stratifying data to prevent performance overestimation, our work relies on KFCV to improve model robustness by training and evaluating multiple subsets of the data.   

\subsection{Transformer-based Planning for Symbolic Regression
} 
In their paper [Shojaee et al.] (2020), they propose a new approach to SR using a Transformer-Based Planning Strategy (TPSR) combined with Monte Carlo Tree Search for equation generation. Their approach improves on traditional symbolic regression methods by integrating non-differentiable feedback into the transformer decoding process. This allows for the generation of equations with better extrapolation abilities compared to classical methods like genetic programming. Similar to our study, [Shojaee et al.] focus on improving and enhancing the generalization and robustness of symbolic regression models. However, the TPSR method improves model performance through the integration of external feedback during the decoding process, while our work ensures better generalization and consistency through k-fold cross-validation. 

\section{Experiments}
For each experiment, we used the 1-variable model. The hyperparameters for the model were set as follows: 20 epochs, a batch size of 128, and an embedding dimension of 512. Computational resources consisted of cloud-based NVIDIA GPUs, specifically three RTX Ada 6000 units, providing a total of 144 GB of VRAM and 186 GB of system RAM. The training and validation speeds were measured at 6 and 11 indices per second, respectively. These specifications were consistent for both the k-fold cross-validation (KFCV) model and the baseline model. Each model was trained from scratch, with the same shortened datasets of 15,000 indices. First is the performance control case, where Figure 2 measures the performance of the model trained on the original 500,000 index dataset using the log of Relative Squared Mean Error (RSME) and Normalized Cumulative Frequency. The figure shows that the model has a high error in a very small quantity of equations and a low error in a large quantity.  
\begin{figure}[b]
  \includegraphics[width=\columnwidth]{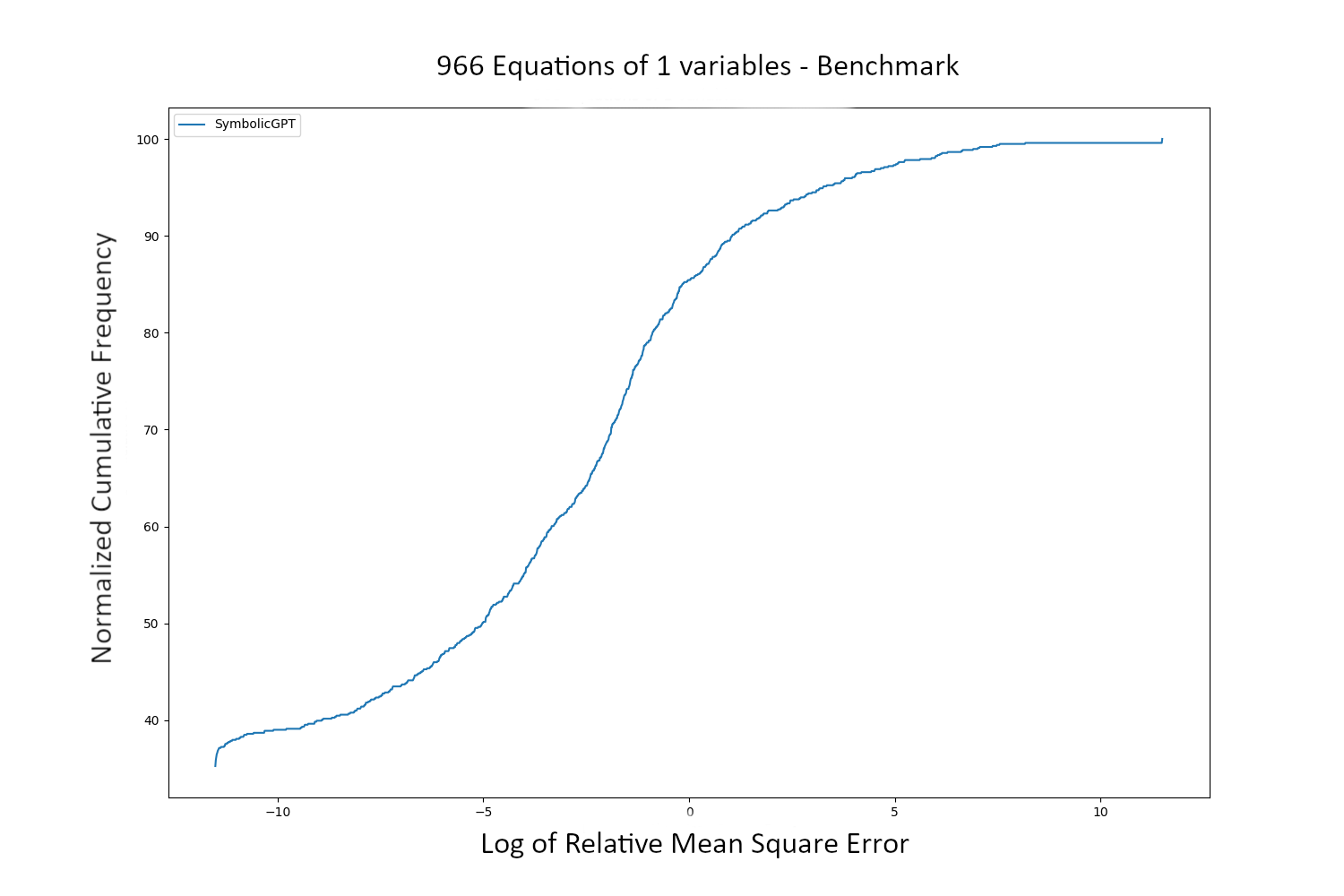
}
  \caption{The following displays the original models performance using log RSME and Normalized Cumulative Frequency}
  \label{fig:experiments4}
\end{figure}

Figure 3 and Table 1 display the learning curve for the old model based on the val loss and train loss metrics, along with a table including the calculated validation loss and train loss per epoch, which averages out to be 0.293467 for train and 0.5966445 for validation.

\begin{figure}[h]
  \includegraphics[width=\columnwidth]{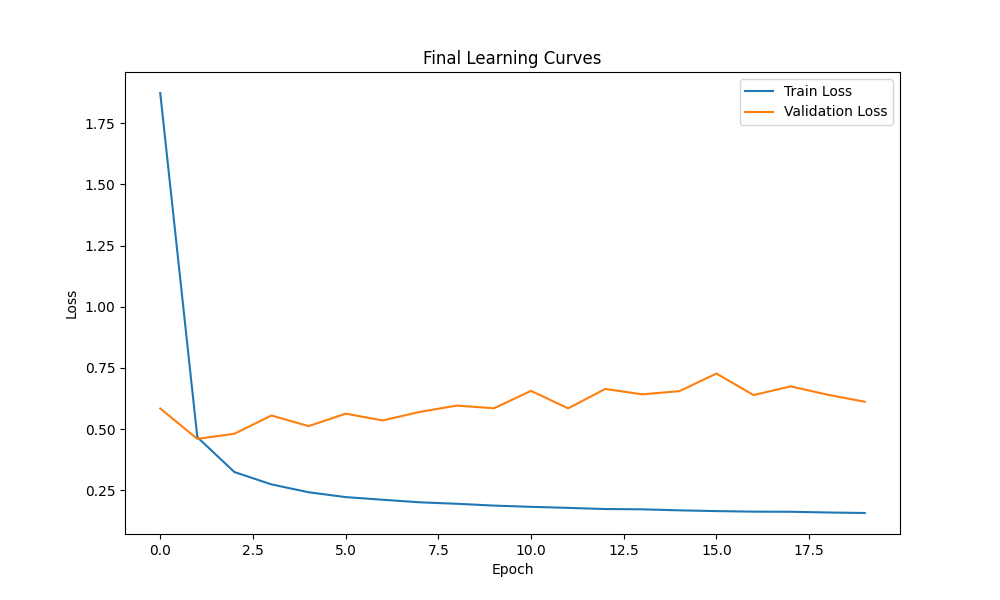}
  \caption{Learning Curve for original model tracking val and train loss per epoch}
  \label{fig:experiments3}
\end{figure}

In comparison, figure 4 and table 2 display the learning curve for the KFCV applied model based on the same metrics, measuring loss per fold rather than per epoch and then averaging that out as well to be an average train loss of 0.26947 and an average validation loss of 0.27858. 

\begin{figure}[b]
  \includegraphics[width=\columnwidth]{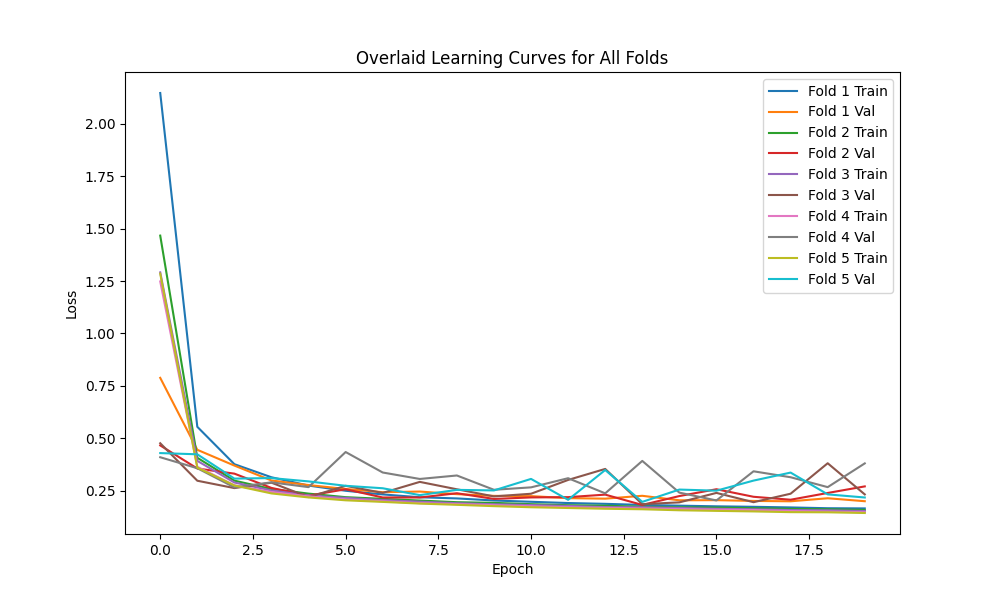}
  \caption{The following graph displays the final learning curves of the KFCV applied model overlaid with all 5 of the folds.}
  \label{fig:experiments1}
\end{figure}

\begin{figure}[h]
  \includegraphics[width=\columnwidth]{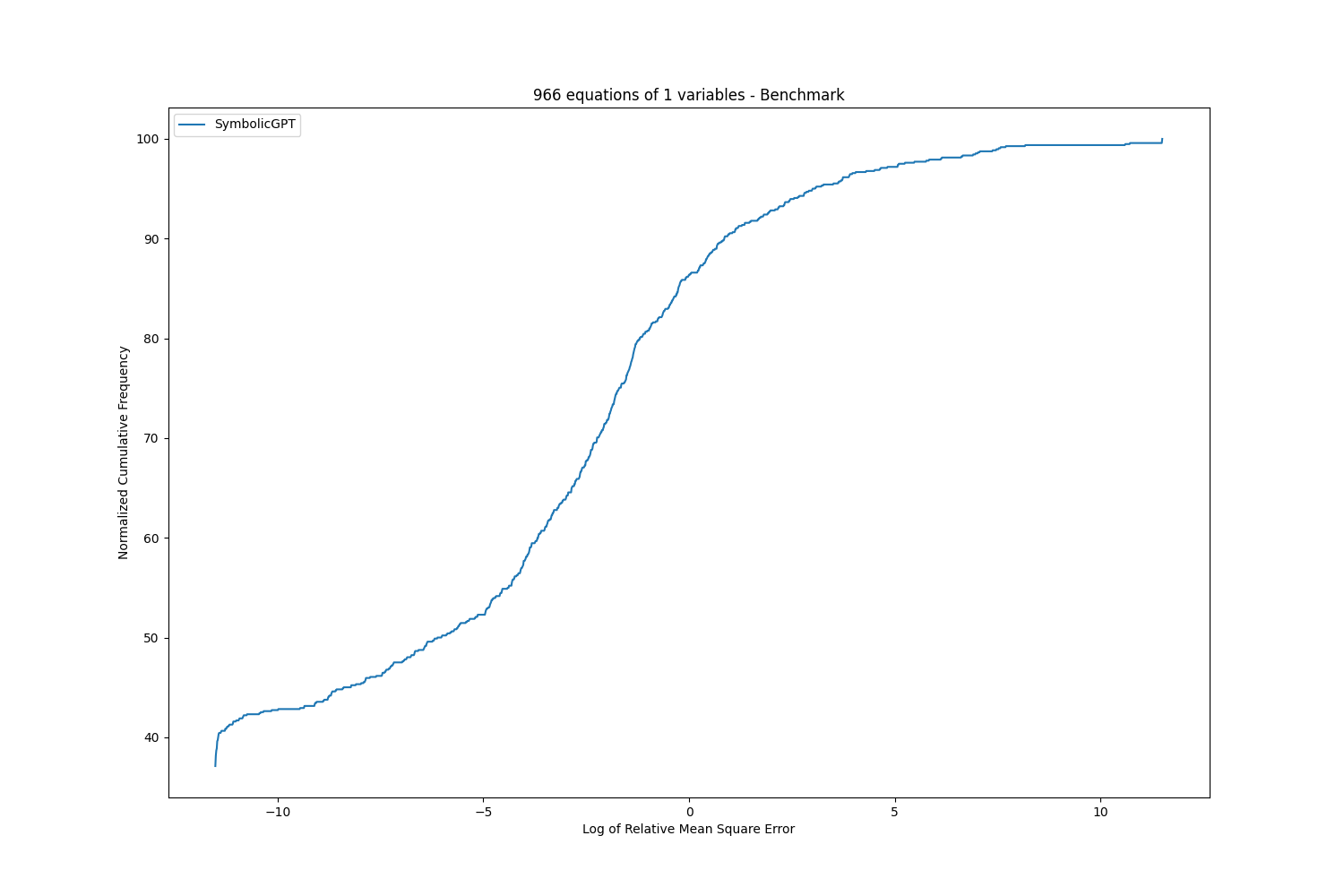}
  \caption{The following displays the KFCV models performance using log RSME and Normalized Cumulative Frequency}
  \label{fig:experiments2}
\end{figure}

Additionally, figure 5 measures the overall performance using the same form and metrics as the base model. It is noticeably nearly identical to the original model trained on a larger dataset. 

\section{Results and Analysis}

Based on the figures and average loss for both the KFCV-applied model and the original model, we calculated the relative validation loss improvement to be 53.31\% using the previously mentioned equation. This demonstrates a significant improvement in generalization in favor of the KFCV-applied model when dealing with smaller datasets. This improvement especially applied to overfitting, which the original model struggled with due to the smaller dataset. Additionally, the overall performance between the 500,000-index model and the 15,000-index model with KFCV was nearly identical. Therefore, we inferred that KFCV impacts the generalizability of SymbolicGPT when trained on small datasets by a substantial margin while retaining strong performance.

\section{Conclusion}

In this paper we have experimented and evaluated the effects of KFCV on SymbolicGPT when trained on smaller datasets, showing a 53.31\% relative generalization improvement when working with a dataset that is 97 percent smaller than usual (500,000 indices to 15,000). Additionally, model performance on the smaller dataset with KFCV remains close to identical to the larger dataset model. This has heavy implications in fields such as physics, finance, and medicine when trying to draw relationships from data that isn't seemingly large enough to train a robust model. KFCV can make up for most of the shortcomings small datasets have while diagnosing any model inefficiencies, such as overfitting. Ultimately, this approach provides a pathway for more reliable and accessible symbolic regression, setting the stage for future innovations in AI and machine learning. 

\section{Limitations}
While our study demonstrates the effectiveness of k-fold cross-validation (KFCV) in improving the generalizability of SymbolicGPT on small datasets, there are several limitations in our experiments that may affect the interpretation of the results. These limitations also highlight potential areas for future investigation.

First of all our experimentation is affected by the limited dataset size variety. Because we only focused on one size, our experimentation may not have captured all of the effects of KFCV. This leaves room for experimentation using larger and smaller datasets to maximize the effects of KFCV.

Secondly, our experimentation focuses on only one model type. Because we only used the 1 variable model, it is possible that the effects of KFCV differ on models with greater numbers of variables. Experimentation could be continued with the inclusion of these models.

Thirdly, We did not take into account for computational overhead. While we reported training speeds, we did not thoroughly explore the additional computational cost of implementing KFCV. This could be significant for larger models or datasets and warrants further investigation.

Lastly, no comparisons were drawn between other methods of small dataset consolidation, such as data augmentation, and KFCV. It would be significant to investigate how KFCV compares to other methods based on efficiency for example.

Regardless of these limitations our study provides valubale insight on the applications of KFCV on Symbolic Regression models trained on small datasets. Investigating the limitations would further advance our understanding of this approach and its applications on the field Symbolic Regression.

\section{Appendix}

\begin{table}[hb]
    \centering
    \begin{tabular}{lcl}
        \hline
        \textbf{Epoch} & \textbf{Train Loss} & \textbf{Validation Loss} \\
        \hline
        1  & 1.87382 & 0.58396 \\
        2  & 0.46677 & 0.45976 \\
        3  & 0.32406 & 0.48057 \\
        4  & 0.27361 & 0.55539 \\
        5  & 0.24153 & 0.51188 \\
        6  & 0.22168 & 0.56281 \\
        7  & 0.21086 & 0.53505 \\
        8  & 0.20035 & 0.56998 \\
        9  & 0.19437 & 0.59583 \\
        10 & 0.18695 & 0.58485 \\
        11 & 0.18185 & 0.65605 \\
        12 & 0.17765 & 0.58455 \\
        13 & 0.17289 & 0.66382 \\
        14 & 0.17167 & 0.64183 \\
        15 & 0.16750 & 0.65485 \\
        16 & 0.16447 & 0.72675 \\
        17 & 0.16218 & 0.63886 \\
        18 & 0.16171 & 0.67435 \\
        19 & 0.15876 & 0.64022 \\
        20 & 0.15666 & 0.61153 \\
        Overall & 0.293467 & 0.5966445 \\

        \hline
    \end{tabular}
    \caption{Train and Validation loss per epoch for original model}
    \label{tab:learning_curve2}
\end{table}

\begin{table}[hb]
    \centering
    \begin{tabular}{lcl}
        \hline
        \textbf{Fold} & \textbf{Train Loss} & \textbf{Validation Loss} \\
        \hline
        1  & 0.32908 & 0.27480 \\
        2  & 0.27029 & 0.25325 \\
        3  & 0.25590 & 0.26972 \\
        4  & 0.24714 & 0.31040 \\
        5  & 0.24494 & 0.28471 \\
        Overall  & 0.26947 & 0.27858 \\

        \hline
    \end{tabular}
    \caption{Train and Validation loss for KFCV applied model}
    \label{tab:learning_curve1}
\end{table}

\bibliography{custom}

\end{document}